\documentclass{llncs}
\usepackage[cmex10]{amsmath}
\usepackage{amssymb}
\usepackage{amsmath}
\usepackage{amsfonts}
\usepackage{algorithm}
\usepackage{algpseudocode}
\usepackage{amsthm}
\usepackage{graphicx}
\usepackage{epsfig}
\usepackage{cite}
\usepackage{tensor}
\usepackage{float}
\usepackage{verbatim}
\usepackage{yhmath}
\graphicspath{{figure/}}
\DeclareGraphicsExtensions{.eps}
\usepackage{gensymb}
\usepackage{subfigure} 
\usepackage{llncsdoc}
\usepackage{color}
\usepackage[dvipsnames]{xcolor}

\begin{document}
\title{Shape Control of Elastic Objects Based on Implicit Sensorimotor Models and Data-Driven Geometric Features}
\author{Wanyu~Ma$^1$, Jihong Zhu$^2$ and David~Navarro-Alarcon$^1$}
\institute{$^1$The Hong Kong Polytechnic University, Hong Kong\\
$^2$TU Delft and Honda Research Institute, Netherlands\\
\email{wanyu.ma@connect.polyu.hk}}
\maketitle

\begin{abstract}
This paper proposes a general approach to design automatic controls to manipulate elastic objects into desired shapes.
The object's geometric model is defined as the shape feature based on the specific task to globally describe the deformation.
Raw visual feedback data is processed using classic regression methods to identify  parameters of data-driven geometric models in real-time.
Our proposed method is able to analytically compute a pose-shape Jacobian matrix based on implicit functions. This model is then used to derive a shape servoing controller.
To validate the proposed method, we report a detailed experimental study with robotic manipulators deforming an elastic rod.
\end{abstract}
\section{Introduction}
The automatic manipulation of rigid objects is one of the canonical problems in robot control; It has been extensively studied in the literature for more than five decades and --- to a great extent --- is considered a solved problem. In recent years, the manipulation of soft deformable objects has attracted the attention of many robotic researchers, mostly due to its multiple potential applications, e.g. palpation of tissues, shaping food materials, handling cables, manipulating fabrics, etc. 

One of the major distinction between manipulation of rigid and deformable objects is the latter’s shape changes during manipulation. Research thus have been focused on controlling the shape of deformable objects. The deformation is often introduced by force on the objects. Therefore, at the very beginning, most researchers preferred to model the deformable object based on accurate physical mechanisms. The most popular methods are the mass-sprint-damping model \cite{kimura2003constructing} and finite element model \cite{petit2017using}, which require to estimate the elastic parameters, such as Young’s modulus and Poisson’s ratio. However, it is impossible to exactly analyze force and deformation and estimate physical parameters for each object to be manipulated since the soft object may be non-homogeneous and the properties at one point of the object could even change over time or the contact with the environment.

Instead of considering force-based modeling, which often requires prior knowledge of the object’s deformation property, later research use vision to perceive the shape as a feedback for deformation control. The area was termed shape servoing \cite{navarro2017fourier}. Shape servoing controls the manipulator for deforming the soft object from the current to the desired shape based on visual feedback.

Image data is often high dimensional hence not directly usable for control. Instead, we represent the shape with feature vectors \cite{chaumette2006visual}. One of the key research question in shape servoing is feature selections. Many features has been proposed for shape control, such as points\cite{wang2018unified}, angles\cite{navarro2016automatic}, curvature\cite{fugl2012simultaneous}, eigenspace \cite{zhu2020visionbased}, catenary\cite{laranjeira2020catenary}, etc. Shape feature with fewer variables improves the controllability of the deformation. Besides, learning or estimating techniques \cite{nair2017combining} can obtain the relationship between deformation and robotic pose. Although these methods don't require physical models, collecting data and training, repeated with the object changing in each specific task, highly increases workload. Hence, this work utilizes continuous geometry, such as curve or surface, to globally and analytically describe the deformable object, and builds up the mapping of deformation and robotic pose based on geometric relationship. The parameters of the geometric mapping are defined as the shape feature which is online identified without prior knowledge of objects’ deformation property.

On the sensing side, most of works mentioned above utilize 2D images \cite{giiler2015estimating}, \ as the feedback. However, 2D feedback will lose one dimension of information resulting in distortion when expressing the real physical world. With the development of the depth camera and the relevant processing methods, the application of 3D data starts to show its advantages. Therefore, more researchers attempt to use 3D data (mostly point cloud) to construct the 3D surface model of the soft object \cite{hu20193}. While, the unorganized raw feedback data from an RGB-D sensor are not able to tell the relationship of one point in the real world with the others, for example, if they belong to one object, or if they are neighbors.  Hence, the classic point cloud processing approach is building topology geometric mesh to simulate and reconstruct their physical relationship \cite{ huang2002combinatorial}.  Although it can obtain a precise surface structure, it results in heavy computing load and redundant information since sometimes the deformation only leads to slight changes in a topological mesh. Therefore, processing data using the topology model is not a good choice to design controllers which require updating the shape very quickly. 

This paper propose a novel framework in shape servoing which makes contributions on both feature selection and 3D sensing:
1) We propose a 3D feature for shape servoing based on a continuous geometric model, and subsequently, using implicit function theorem to obtain an analytical Jacobian matrix; 
2) On the sensing side, we process high dimensional feedback in real-time by dealing with unorganized raw visual feedback with less computational demanding.

The rest of this paper is organized as follows. Section 2 presents a general approach. Section 3 elaborates the mechanism of the proposed approach based on a specific case. Section 4 analyses and verifies our approach by robotic experiments. Finally, the conclusion is given in Section 5.

\section{Methodology}
In this section, an overview of the framework is given at first. Then, the detailed skills are later elaborated.
\subsection{Overview}
First, we determine the research target, that is, the type of soft object, the feedback data from the sensor, the desired result, etc. Then, according to this target, select a suitable geometric model to approximate the shape of the object. For example, when we study a linear deformable object, we could use a conic section on a spatial plane if the soft object is bent and use a helix if the soft object is twisted. 

Second, we analyze the geometric relationship between the soft object and the pose of the robotic manipulator relying on the selected shape feature. Then, compute the analytical pose-shape Jacobian matrix by taking the partial derivatives of the pose of the manipulator with respect to the shape feature parameters. Since implicit functions are the most likely used to describe the geometric relationship, the implicit function theorem is introduced to get the pose-shape Jacobian matrix. 

Finally, we design a velocity-based controller in the task space using the obtained analytical pose-shape Jacobian matrix, for this matrix describes the velocity relationship. 

After the analysis is done, the control law is able to be used to update the control command for the manipulator in each loop. Once the control process starts, the raw data should be online fitted to the selected geometric model to identify shape feature parameters. This is a typical regression problem, which could be either linear or nonlinear. The frequent solutions are the Least Square Method (LSM), Gradient Descent, Newton's method, Quasi-Newton Methods, Conjugate Gradient, etc. 
\subsection{Online Identification of Shape Feature}
In this section, the regression method is used to fit raw data to a data-driven geometric model identifying shape features in every control loop. We choose LSM for it is simple and fast. LSM\cite{axelsson1987generalized} is a classical linear regression algorithm whose core is to identify the parameters of the model to minimize the sum of squared residual defined as the difference between the observed value by sample and the predicted value by model.

Let $\boldsymbol{S}=[\boldsymbol{s}_1,\cdots,\boldsymbol{s}_N] \in \mathbb{R}^{n \times N}$ denote the unorganized raw feedback from sensors where $\boldsymbol{s}_i$ is a single $n$-dimensional element of data. 
Assume there are $N$ elements in set $\boldsymbol{S}$ used for parameter identification.
Denote $m$-dimensional parameters vector of shape feature as $\boldsymbol{y} \in \mathbb{R}^m$ which is online identified during the control process. 
The mapping between feedback $\boldsymbol{s}$ and shape feature parameter $\boldsymbol{y}$ satisfies $\boldsymbol{f}\left([\boldsymbol{s}^\text{T},\boldsymbol{y}^\text{T}]^\text{T}\right)=\boldsymbol{0}_{l}$,
where $\boldsymbol{f}$ is an implicit function set comprising $l$ functions $\boldsymbol{f}=\{f_1,\cdots,f_l\}$, and $f_i (i=1 \cdots l)$ is a twice continuously differentiable function.
The fitting process is an unconstrained optimization problem 
\begin{equation}
    \min_{\boldsymbol{y}} \left\{\sum^l_{i=1} f_i^2([\boldsymbol{s}^\text{T},\boldsymbol{y}^\text{T}]^\text{T}):\boldsymbol{s} \in \mathbb{R}^n, \boldsymbol{y} \in \mathbb{R}^m \right\}.
\label{eq:rasidual}
\end{equation}
\subsection{Derivation of Analytical Jacobian Matrix}
Jacobian matrix is a tool to describe the velocity relationship which is obtained from the first-order partial derivatives of the displacement mapping. Traditionally, the mapping should be explicit, while it is hard to obtain for the selected geometric model. Hence, to derive a Jacobian matrix based on implicit mapping, we firstly introduce implicit function theorem\cite{jittorntrum1978implicit, ha2017joint} as follows:
\begin{lemma}[Implicit function theorem]
Let $\boldsymbol{h}\left([\boldsymbol{x}^\text{T},\boldsymbol{y}^\text{T}]^\text{T}\right)=\boldsymbol{0}:\mathbb {R} ^{q+m}\to \mathbb {R} ^m$ be a continuously differentiable function of two sets of variables, $\boldsymbol{x}\in \mathbb{R}^q$ and $\boldsymbol{y}\in\mathbb{R}^m$. 
If the Jacobian matrix $\boldsymbol{J}_{\boldsymbol{h},\boldsymbol{x}} = \frac{\partial\boldsymbol{h}}{\partial\boldsymbol{x}}\in\mathbb{R}^{(q+m)\times q}$
is invertible, the Jacobian matrix of $\boldsymbol{x}$ with respect to $\boldsymbol{y}$ is given by the matrix product
\begin{equation}
    \boldsymbol{J}_{\boldsymbol{x},\boldsymbol{y}} =\frac{\partial\boldsymbol{x}}{\partial\boldsymbol{y}}=-\left[\frac{\partial\boldsymbol{h}}{\partial\boldsymbol{x}}\right]^{-1}  \frac{\partial\boldsymbol{h}}{\partial\boldsymbol{y}} \in\mathbb{R}^{q\times m}
\end{equation}
\label{lemma1}
\end{lemma}

Then, let $\boldsymbol{x}\in\mathbb{R}^q$ denote the pose of the robot end-effector, which is the feedback from the robotic manipulator. The analytical pose-shape Jacobian matrix $\boldsymbol{J}_S \in \mathbb{R}^{q \times m}$ is defined as:
\begin{equation}
\dot{\boldsymbol{x}}=
    \boldsymbol{J}_S\dot{\boldsymbol{y}}
\label{eq:jacobian}
\end{equation}
To obtain the Jacobian matrix $\boldsymbol{J}_{S}$, geometric relation between shape feature $\boldsymbol{y}$ and the pose of the end-effector $\boldsymbol{x}$ should be developed: 
\begin{equation}
    \boldsymbol{h}\left([\boldsymbol{y}^\text{T},\boldsymbol{x}^\text{T}]^\text{T}\right)=\boldsymbol{0}_{p}
    \label{mapping}
\end{equation}
and $\boldsymbol{h}$ has $p$ functions $\boldsymbol{h}=\{ h_{1},\cdots,h_{p}\}$ where each $h_{i}(i=1\cdots p)$ is continuously differentiable, probably either a linear function or a nonlinear function.
Then, we make derivative of Eq.(\ref{mapping}):
\begin{equation}
    \boldsymbol{h}'=
    \left[
    \begin{matrix}
    \boldsymbol{J}_1 & \boldsymbol{J}_2
    \end{matrix}
    \right]
    \left[
    \begin{matrix}
    \dot{\boldsymbol{y}}\\
    \dot{\boldsymbol{x}}
    \end{matrix}
    \right]
    =\boldsymbol{0}_{p},\ 
    \text{where\ }
    \label{15}
\boldsymbol{J}_1=\frac{\partial\boldsymbol{h}}{\partial\boldsymbol{y}}\in\mathbb{R}^{p\times m},\ \ 
\boldsymbol{J}_2=\frac{\partial\boldsymbol{h}}{\partial\boldsymbol{x}}\in\mathbb{R}^{p\times q}
\end{equation}
According to Eq.(\ref{15}) and Lemma \ref{lemma1}, we yield $\boldsymbol{J}_1\dot{\boldsymbol{y}}+\boldsymbol{J}_2\dot{\boldsymbol{x}}=\boldsymbol{0}_{p} \Rightarrow \dot{\boldsymbol{x}}=-\boldsymbol{J}_2^\dagger\boldsymbol{J}_1\dot{\boldsymbol{y}}$.
Therefore, we obtain the Jacobian matrix $\boldsymbol{J}_S$ as follows:
\begin{equation}
    \boldsymbol{J}_{S}=-\boldsymbol{J}_2^\dagger\boldsymbol{J}_1
\end{equation}
Note that, in this paper, we use pseudoinverse trick to take inverse of a matrix since the matrix is probably not a full-rank matrix. 
\subsection{Designing Controller}
The desired shape feature is denoted as $\boldsymbol{y}_d$. The proposed shape servoing controller aims at minimizing $\boldsymbol{e} = \boldsymbol{y} - \boldsymbol{y}_d$ through robotic motion \cite{hutchinson2006visual,chaumette2007visual}. 
Given the desired shape $\boldsymbol{y}_d$ and its deformation velocity $\dot{\boldsymbol{y}}_d$, a task space controller can be designed based on analytical Jacobian matrix $\boldsymbol{J}_S$ and online updating shape-feature parameters $\boldsymbol{y}$.
The desired velocity of the 
manipulator end-effector can be obtained by
\begin{equation}
\dot{\boldsymbol{x}}=\boldsymbol{J}_S\left(\dot{\boldsymbol{y}}_d-\boldsymbol{K}(\boldsymbol{y}-\boldsymbol{y}_d)\right)
\label{dx}
\end{equation}
where $\boldsymbol{K}$ is a positive definite gain matrix. The desired end-effector trajectory $\boldsymbol{x}$ is obtained by numerical integration.

To proof the stability, we select a Lyapunov function $V= \frac{1}{2} \boldsymbol{e}^\text{T} \boldsymbol{e}$,
where $\boldsymbol{e}=\boldsymbol{y}-\boldsymbol{y}_d$. 
Take the derivative of $V$ to yield $\dot{V}=\boldsymbol{e}^\text{T}\dot{\boldsymbol{e}}$,
where $\dot{\boldsymbol{e}}=\dot{\boldsymbol{y}}-\dot{\boldsymbol{y}}_d$. 
Submit Eq.(\ref{dx}) in it, we yield $\dot{\boldsymbol{e}}=\boldsymbol{J}_S^\dagger\dot{\boldsymbol{x}}-[\boldsymbol{J}_S^\dagger\dot{\boldsymbol{x}}+\boldsymbol{K}(\boldsymbol{y}-\boldsymbol{y}_d)]
    =-\boldsymbol{K}\boldsymbol{e}$.
Then, submit this $\dot{\boldsymbol{e}}$ into $\dot{V}$, the derivative becomes $\dot{V}=-\boldsymbol{e}^\text{T}\boldsymbol{K}\boldsymbol{e}\leq 0$.
Finally, the global stability of the controller is proofed.
\section{Case of Study: Bending Based on Spatial Arc}
To demonstrate how the general approach works, a specific deformable object manipulation task is going to be solved under the proposed framework. 

\subsection{Analysis of the Task}
\begin{figure}[htb]
\vspace{-0.7cm} 
\setlength{\abovecaptionskip}{-0.1cm}  
\setlength{\belowcaptionskip}{-0.8cm} 
    \centering
    \includegraphics[width=0.5\linewidth]{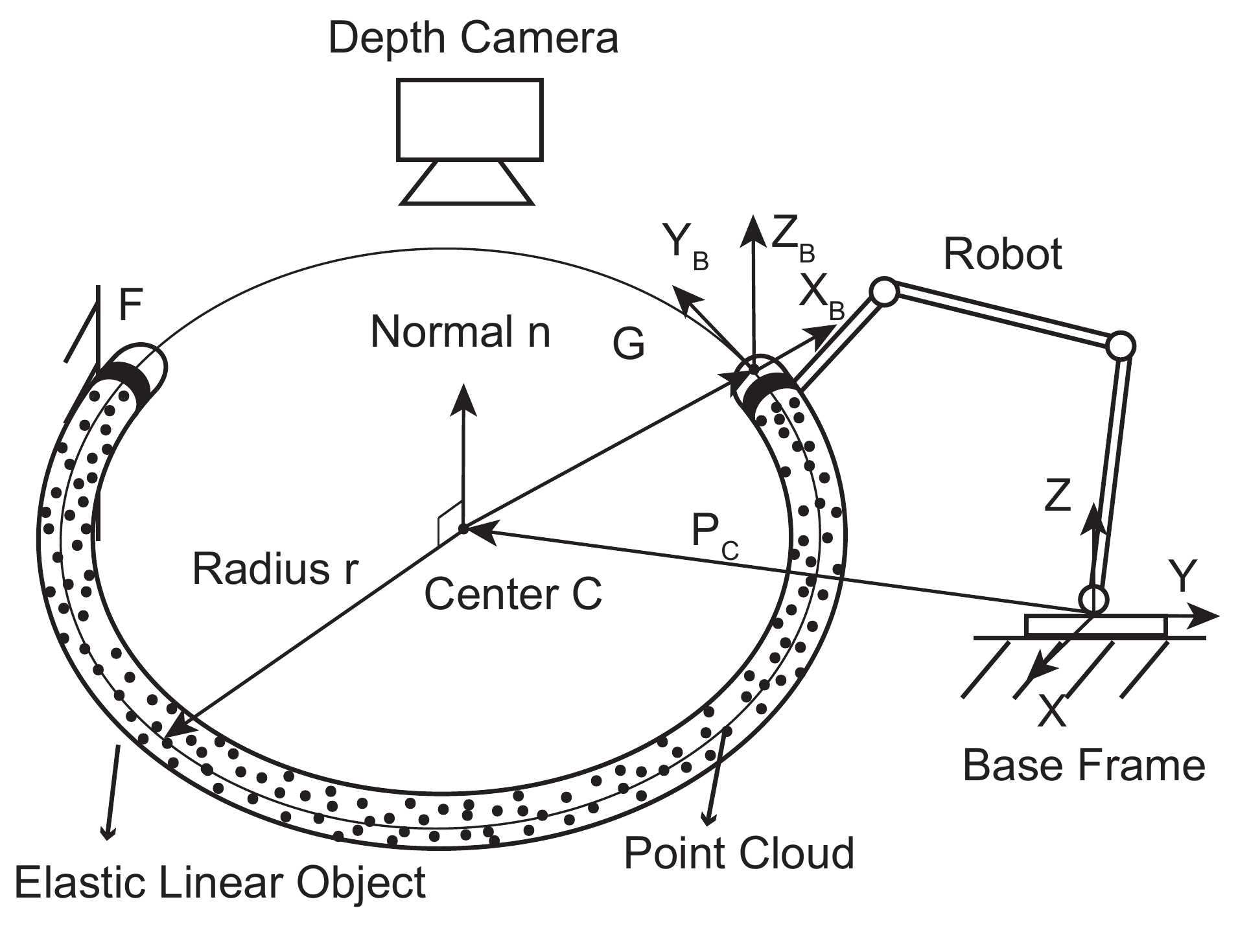}
    \caption{Conceptual illustration of the setup and the shape feature (spatial arc).}\hfill
    \label{fig:setup+circle}
\end{figure}

We select a linear elastic deformable object as the research target. A vision-based shape servoing controller will be designed to bend the object into the desired curvature and position defined by a continuous geometric curve. 
The setup consisting of an RGB-D camera and a robotic arm will be used to conduct the experiment, of which, the conceptual demonstration is shown in Fig.\ref{fig:setup+circle}.

For the deformable object, one end tip is fixed at a known point $F$ and the other tip is grasped at point $G$ by a robot arm without relative motion. 
Both $F$ and $G$ are on the surface of the deformable object.
The feedback provided by the robotic manipulator includes three-axial Cartesian coordinates and Euler angles under the rotation sequence Z-Y-Z, which describe respectively position and orientation of the end-effector, that is, $q=6$. 
Denote Euler angles as $\left[ x_1, x_2, x_3 \right]^\text{T}$ and Cartesian position as $\left[ x_4, x_5, x_6 \right]^\text{T}$, forming 6D robotic pose $\boldsymbol{x} = \left[ x_1, x_2, x_3, x_4, x_5, x_6\right]^\text{T}$. 
It is assumed that the deformable object in this work has elastic performance maintaining its shape against the manipulation of the robot arm, and gravity, occlusion, overlapping, crossing, twisting are not considered. 

Here gives some denotations. The vector $\boldsymbol{p}_A=[x_A,y_A,z_A]^\text{T}$ represents the position of point $A$ with respect to the robot base frame. 
The vector $\boldsymbol{d}_A = [d_{Ax}, d_{Ay}, d_{Az}]^\text{T} = \boldsymbol{p}_G - \boldsymbol{p}_A$ represents the vector from point $A$ to the robot end-effector, that is, the grasped point $G$ in this paper. 

The selected shape feature is spatial arc, whose description is a part of a spatial circle composed of radius, center, and normal vector of the plane where the circle is, respectively denoted as $r$, $\boldsymbol{p}_C = [x_C, y_C, z_C]^\text{T}$, and $\boldsymbol{n}=[n_x, n_y, n_z]^\text{T}$. Hence, the shape feature is reformed as a parameter vector $\boldsymbol{y}=[r, \boldsymbol{p}_C^\text{T}, \boldsymbol{n}^\text{T}]^\text{T} \in \mathbb{R}^7$.
The normal vector is limited to a unit vector.
Fig.\ref{fig:setup+circle} demonstrates the component of a 3D circle and how the spatial conic section represents the deformable object based on the setup.

\subsection{Analytical Jacobian Matrix}
We separately analyze the orientation Jacobian matrix and position Jacobian matrix when we compute analytical Jacobian matrix $\boldsymbol{J}_S$ between the deformation velocity $\dot{\boldsymbol{y}}$ and the end-effector velocity $\dot{\boldsymbol{x}}$. It is reasonable because we usually decouple the geometric relationship from orientation and position.
The orientation Jacobian matrix and position Jacobian matrix are denoted by $\boldsymbol{J}_{SO}$ and $\boldsymbol{J}_{SP}$, forming the pose-shape Jacobian matrix $\boldsymbol{J}_S = \left[ \boldsymbol{J}_{SO}^\text{T}, \boldsymbol{J}_{SP}^\text{T} \right]^\text{T}$.

To derive the mapping of pose and deformation in terms of orientation, we define two frames, namely, the body frame of the deformable object and the end-effector frame, respectively denoted as $\boldsymbol{F}_B$ and $\boldsymbol{F}_E$. 
Corresponding rotation matrices $\boldsymbol{R}_B$ is computed by shape features $\boldsymbol{y}$, and $\boldsymbol{R}_E$ is computed by robot pose $\boldsymbol{x}$. 
The body frame is represented by $\{\boldsymbol{F}_B\}=\{G, \boldsymbol{X}_B, \boldsymbol{Y}_B, \boldsymbol{Z}_B\}$.

Since there is no relative motion between the end-effector and the grasped object, we assume that $\boldsymbol{F}_B$ has a constant transformation with respect to $\boldsymbol{F}_E$, namely, $\boldsymbol{R}_E={}^E\!\boldsymbol{R}_B \boldsymbol{R}_B $, where the constant orientation relationship ${}^E\!\boldsymbol{R}_B$ can be initialized by ${}^E\!\boldsymbol{R}_B = \boldsymbol{R}_{E0} \boldsymbol{R}_{B0}^\text{T}$. 

To describe the orientation of the deformable object, the body frame $\{\boldsymbol{F}_B\}$ should be defined, where origin is the grasping point $G$, axis $\boldsymbol{X_B}$ is a unit vector along vector $\boldsymbol{d}_C$, axis $\boldsymbol{Z_B}$ is a unit vector along the normal $\boldsymbol{n}$, and axis $\boldsymbol{Y_B}$ is determined by right-hand rule. That is, $\boldsymbol{X}_B=\frac{\boldsymbol{d}_C}{\| \boldsymbol{d}_C \|}, \boldsymbol{Z}_B=\boldsymbol{n}, \boldsymbol{Y}_B=\boldsymbol{Z}_B \times \boldsymbol{X}_B$.
The orientation of the body frame $\{\boldsymbol{F}_B\}$ is described by rotation matrix $\boldsymbol{R_B} =  \left[ \frac{\boldsymbol{d}_C}{\| \boldsymbol{d}_C \|}, \frac{\boldsymbol{n} \times \boldsymbol{d}_C}{\| \boldsymbol{d}_C \|}, \boldsymbol{n} \right]$.
The rotation matrix $\boldsymbol{R}_E$ can be obtained by Euler angles under the rotation sequence Z-Y-Z:
\begin{equation}
\begin{aligned}
    \boldsymbol{R}_E  =\boldsymbol{R}_Z(x_{1})\boldsymbol{R}_Y(x_{2})\boldsymbol{R}_Z(x_{3})
    = \left[
 \begin{matrix}
    c_1c_2c_3-s_1s_3\ \  & -c_1c_2s_3-s_1c_3\ \  & c_1s_2\\
    s_1c_2c_3+c_1s_3\ \  & -s_1c_2s_3+c_1c_3\ \  & s_1s_2\\
    -s_2c_3              & s_2s_3                & c_2\\
  \end{matrix}
  \right]
\end{aligned}
\label{RE}
\end{equation}
and $s_1$, $c_1$, $s_2$, $c_2$, $s_3$, $c_3$ represent respectively $\sin{x_{1}}$, $\cos{x_{1}}$, $\sin{x_{2}}$, $\cos{x_{2}}$, $\sin{x_{3}}$, $\cos{x_{3}}$.
Reshaping $\boldsymbol{X}_B$ and $\boldsymbol{Z}_B$ into a vector with approximating $\| \boldsymbol{d}_C \|=r$ yields $\boldsymbol{V}_1 = \left[\frac{\boldsymbol{d}_{C}}{r}^\text{T}, \boldsymbol{n}^\text{T} \right]^\text{T}$.
Reshaping the corresponding $X$ axis and $Z$ axis of $\boldsymbol{R}_E$ into a vector yields $\boldsymbol{V}_2 = \left[c_1c_2c_3-s_1s_3, s_1c_2c_3+c_1s_3, -s_2c_3, c_1s_2, s_1s_2, c_2 \right]^\text{T}$. 
Then, the orientation mapping is: 
\begin{equation}
    \boldsymbol{h}_O =
     {}^E\boldsymbol{T}_B
  \boldsymbol{V}_1-\boldsymbol{V}_2
    =\boldsymbol{0},~
    {}^E\boldsymbol{T}_B  = 
    \left[
 \begin{matrix}
  {}^E\boldsymbol{R}_B & \boldsymbol{0}_{3 \times 3} \\
  \boldsymbol{0}_{3 \times 3} & {}^E\boldsymbol{R}_B
  \end{matrix}
  \right] 
\end{equation}
Therefore, the orientation Jacobian matrix is $\boldsymbol{J}_{SO}=-\boldsymbol{J}_2^\dagger \boldsymbol{J}_1 \in\mathbb{R}^{3\times 7}$, 
where the Jacobian matrices $\boldsymbol{J}_1={}^E\boldsymbol{T}_B \tilde{\boldsymbol{J}}_1$ and $\boldsymbol{J}_2$ are:
\begin{equation}
\small
\tilde{\boldsymbol{J}}_1=
 \left[
 \begin{matrix}
   -\frac{d_{Cx}}{r^2} & -\frac{1}{r} & 0 & 0 & 0 & 0 &0\\
   -\frac{d_{Cy}}{r^2} & 0 & -\frac{1}{r} & 0 & 0 & 0 &0\\
   -\frac{d_{Cz}}{r^2} & 0 & 0 & -\frac{1}{r} & 0 & 0 &0\\
   0 & 0 & 0 & 0 & 1 & 0 & 0\\
   0 & 0 & 0 & 0 & 0 & 1 & 0\\
   0 & 0 & 0 & 0 & 0 & 0 & 1
  \end{matrix}
  \right],~
\boldsymbol{J}_2=
 \left[
 \begin{matrix}  
   s_1c_2c_3+c_1s_3 & c_1s_2c_3 & c_1c_2s_3+s_1c_3\\
   -c_1c_2c_3+s_1s_3 & s_1s_2c_3 & s_1c_2s_3-c_1c_3\\
   0 & c_2c_3 & -s_2s_3\\
   s_1s_2 & -c_1c_2 & 0\\
   -c_1s_2 & -s_1c_2 & 0\\
   0 & s_2 & 0
  \end{matrix}
  \right]
\label{J2}
\end{equation}

Since there is no relative motion between the end-effector and the grasped object, we assume that the position of the grasped point is equal to the feedback position of the robot, namely,  $\boldsymbol{p}_G = \left[ x_4, x_5, x_6 \right]^\text{T}$.

According to the geometric information, the relationship between end-effector position and the shape feature $\boldsymbol{h}_P=\boldsymbol{0}$ can be described as follows:
\begin{equation}
\boldsymbol{h}_P=
\left\{
 \begin{array}{l}
    \arccos{\frac{\overrightarrow{CF} \cdot \boldsymbol{d}_C}{r^2}}-\theta  = 0\\
    \boldsymbol{n} \cdot \boldsymbol{d}_C  = 0\\
    \boldsymbol{d}_C \cdot \boldsymbol{d}_C-r^2   = 0
\end{array}
\right.
\end{equation}
in which, $\theta$ is the angle swiping from $\overrightarrow{CF}$ to $\boldsymbol{d}_C$ through the object.

\begin{figure}[h!]
\vspace{-0.7cm} 
\setlength{\abovecaptionskip}{-0.2cm}  
\setlength{\belowcaptionskip}{-0.6cm} 
    \centering
    \subfigure[Case2]{
    \label{fig:detect-circle-arc-length-1}
    \includegraphics[width=0.3\linewidth]{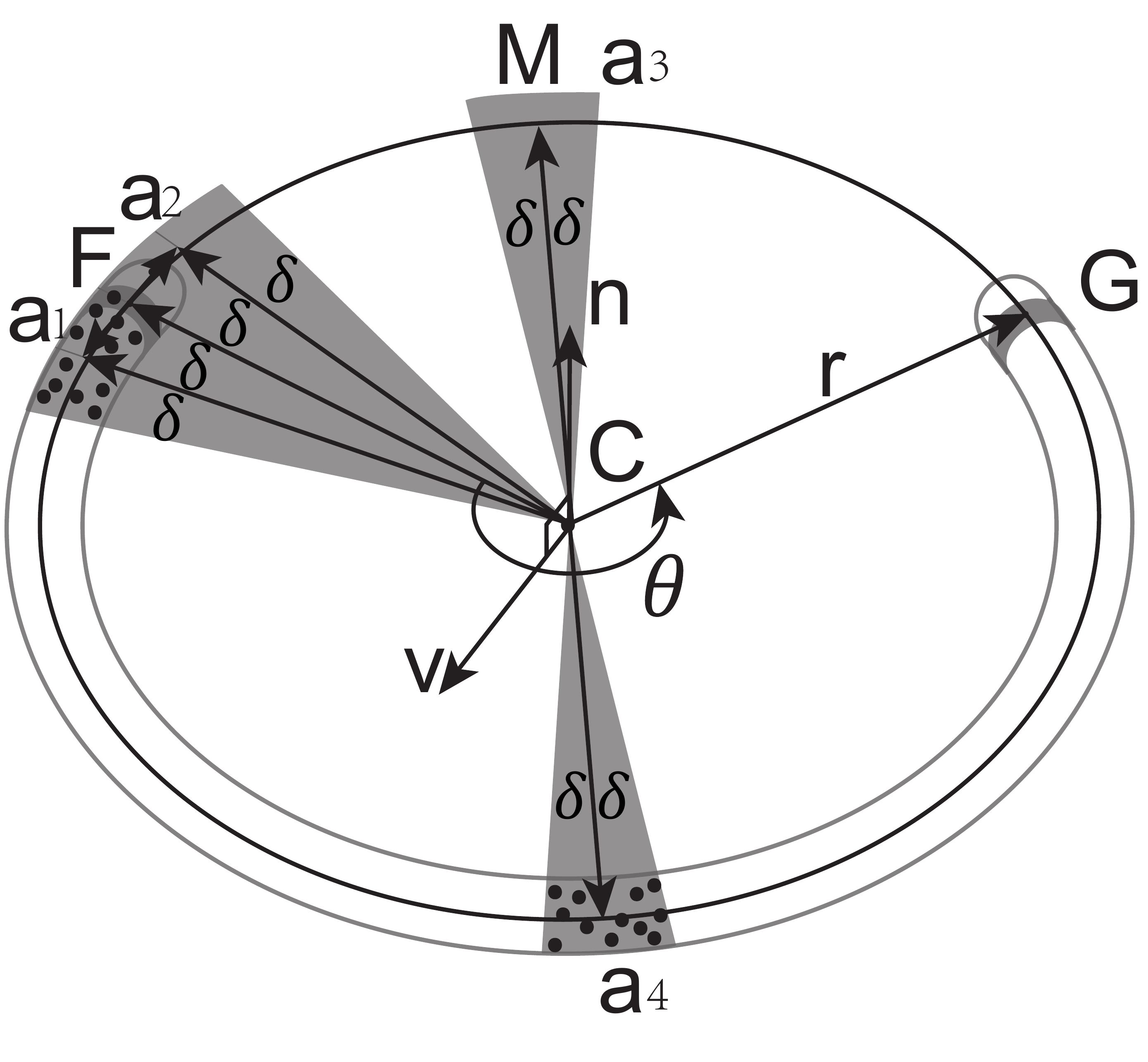}
}\ \ \ \ \ \ \ \ \ \ \ \ 
    \subfigure[Case3]{
    \label{fig:detect-circle-arc-length-2}
    \includegraphics[width=0.3\linewidth]{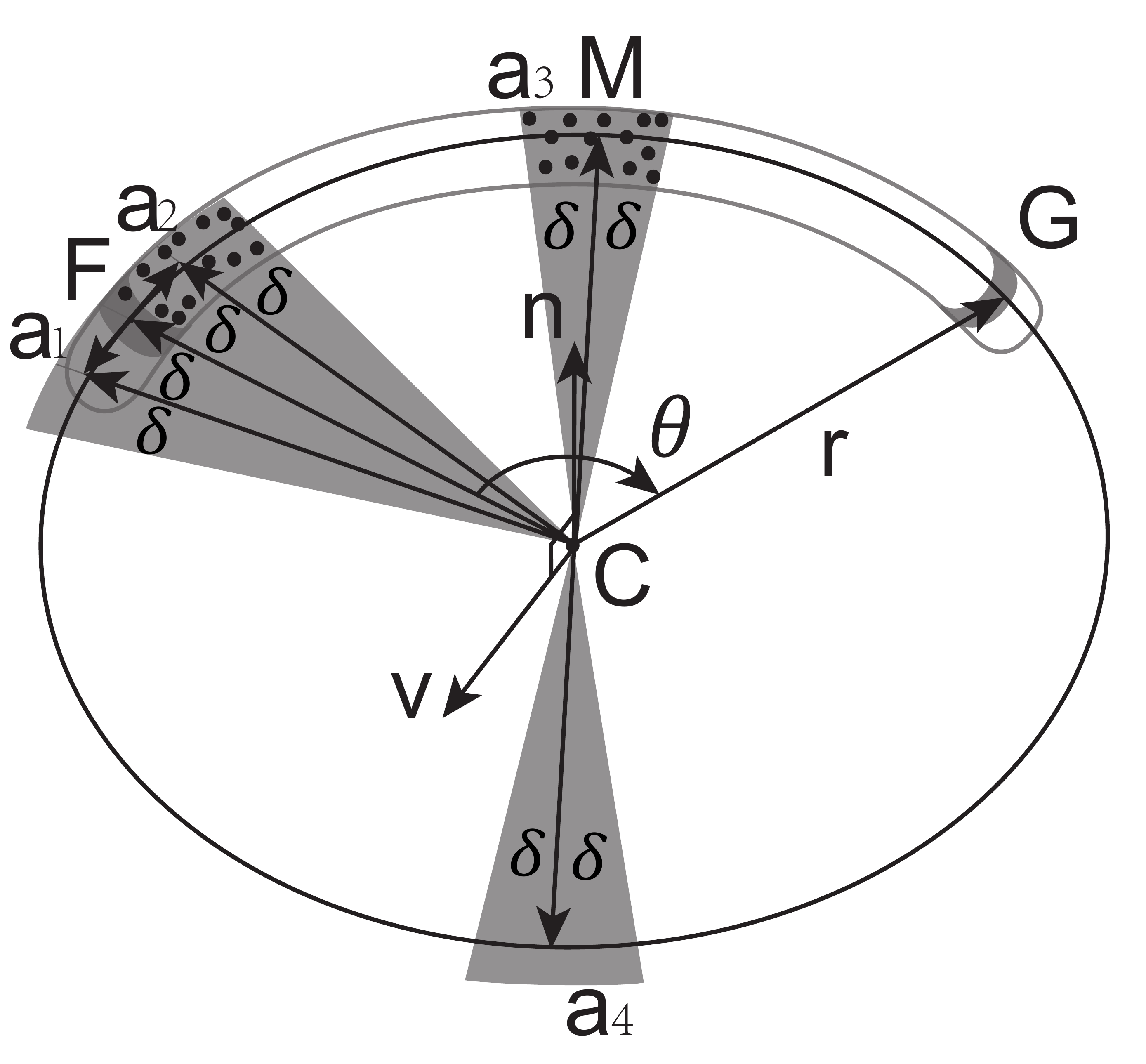}
}
    \caption{Two examples of angle $\theta$ configuration.}
    \label{fig:detect-circle-arc-length} 
\end{figure}
Since the length of the manipulated deformable object is constant, the arc length $L$ of the identified curve doesn't change either, which should limit and guide the robot moving. 
For example, as the configuration is shown in Fig.\ref{fig:detect-circle-arc-length-1}, if we want the radius decreasing, the robot should move counterclockwise while if the configuration is like Fig.\ref{fig:detect-circle-arc-length-2}, the robot should clockwise move to decrease the radius, which means, though the angle $\theta$ is the same in two configurations, the robot motion is different.
In this way, it is necessary to determine the exact geometric relationship under the limitation of the arc length.

Firstly, note that, since the setup is predesignated, the manipulated object has a limited and fixed workspace. 
Thus, during the process of manipulation, we can set the normal vector $\boldsymbol{n}$ always pointing to one side of the plane, though two vectors along opposite directions are both able to describe a plane. 
Then, we define a vector $\boldsymbol{v} = \boldsymbol{n} \times \overrightarrow{CF}$ and a small positive angle $\delta$.
$\boldsymbol{a}_1 = \overrightarrow{CF} + \delta \boldsymbol{v}, \boldsymbol{a}_2 = \overrightarrow{CF} - \delta \boldsymbol{v}$ approximately represent the vectors forming a small angle $\delta$ with $\overrightarrow{CF}$ counterclockwise and clockwise.
$\boldsymbol{a}_3 = \overrightarrow{CF} + \boldsymbol{d}_C$ is the midline of the acute angle between $\overrightarrow{CF}$ and $\boldsymbol{d}_C$, and $\boldsymbol{a}_4 = -\boldsymbol{a}_3$ is the midline of the obtuse angle.
Surrounding $\boldsymbol{a}_1, \boldsymbol{a}_2, \boldsymbol{a}_3, \boldsymbol{a}_4$, generate four areas of feedback points. 
All vectors, starting from the center $C$ pointing to the feedback point in the area, form an angle less than $\delta$ with $\boldsymbol{a}_1, \boldsymbol{a}_2, \boldsymbol{a}_3, \boldsymbol{a}_4$ respectively (as the shadows in Fig.\ref{fig:detect-circle-arc-length}). 
Then, denote the number of the feedback points which is within the defined areas using $N_1, N_2, N_3, N_4$, by computing the angles between the vectors. 
This detection technique to determine $\theta$ and initial $L$ is demonstrated by Table \ref{table:detect}.
$r_0$, $C_0$, $G_0$, and $F_0$ denote initial states of radius, center, grasped point, and fixed point.
\begin{table}[ht!]
\vspace{-0.7cm} 
\setlength{\belowcaptionskip}{-0.7cm} 
\centering
\begin{tabular}{c| c | c | c | c} 
 \hline
  & $N_1 \& N_2$ & $N_3 \& N_4$ & $\theta$ & $L$ \\ 
 \hline
  Case1 & $N_1 > N_2$ & $N_3 > N_4$ & $\frac{L}{r}$         & $r_0 \cdot \arccos{\frac{\overrightarrow{C_0 F_0} \cdot \overrightarrow{C_0 G_0}}{r_0^2}}$\\ 
  Case2 & $N_1 > N_2$ & $N_3 < N_4$ & $2\pi - \frac{L}{r }$ & $2\pi r_0 - r_0 \cdot \arccos{\frac{\overrightarrow{C_0 F_0} \cdot \overrightarrow{C_0 G_0}}{r_0^2}}$ \\
  Case3 & $N_1 < N_2$ & $N_3 > N_4$ & $-\frac{L}{r}$        & $r_0 \cdot \arccos{\frac{\overrightarrow{C_0 F_0} \cdot \overrightarrow{C_0 G_0}}{r_0^2}}$\\ 
  Case4 & $N_1 < N_2$ & $N_3 < N_4$ & $-2\pi + \frac{L}{r}$ & $2\pi r_0 - r_0 \cdot \arccos{\frac{\overrightarrow{C_0 F_0} \cdot \overrightarrow{C_0 G_0}}{r_0^2}}$ \\
 \hline
\end{tabular}
\caption{Detection technique to determine $\theta$ and initialize $L$}
\label{table:detect}
\end{table}

Then, we derive the pose-shape Jacobian matrix for position $\boldsymbol{J}_{SP}=-\boldsymbol{J}_2^\dagger \boldsymbol{J}_1 \in\mathbb{R}^{3\times 7}$.
Since there are four cases in Table \ref{table:detect}, $\boldsymbol{J}_2$ and $\boldsymbol{J}_1$ also have four cases respectively. 
Here we give the solution for Case2 by computing the derivatives of $\boldsymbol{h}_P$ (the other three cases should have similar solutions):
\begin{equation}
\boldsymbol{J}_1=
 \left[
 \begin{matrix}
  \frac{-2r\gamma \eta + L}{r^2} & -\gamma \left(\overrightarrow{CF}^\text{T}+\boldsymbol{d}_C^\text{T} \right) & \boldsymbol{0_{1 \times 3}}\\
  0 & -\boldsymbol{n}^\text{T} & \boldsymbol{d}_C^\text{T}\\
  -2r & -2\boldsymbol{d}_C^\text{T} & \boldsymbol{0_{1 \times 3}}
  \end{matrix}
  \right],~
\boldsymbol{J}_2=
 \left[
 \begin{matrix}
  \gamma \overrightarrow{CF} & \boldsymbol{n} & 2\boldsymbol{d}_C
  \end{matrix}
  \right]^\text{T}
\end{equation}
where $\gamma = \frac{-1}{\sqrt{1- \eta^2}}$, and $\eta = \frac{\overrightarrow{CF} \cdot \boldsymbol{d}_C}{r^2}$.

\section{Experiments and Results}
In this section, we conduct experiments to validate our approach. 
We aim to achieve a converging controller to control the manipulator bend of a linear elastic soft object (elastic rod) into the desired shape. 
The real setup consists of an Intel RealSense D415 depth camera and a UR3 robot arm, where the depth camera is settled with the system in an eye-to-hand manner. 
The calibration is needed before the experiment. 

Firstly, to validate the algorithm of shape feature identification, four point-clouds are collected under the different configurations of the manipulated object.
The collected data is 10 times downsampled so that every point cloud has about 200 points.
A filter is used to reduce the noise of raw visual feedback. 
Then, LSM is used to compute the shape feature parameters.

\begin{figure}[ht!]
\vspace{-0.7cm} 
\setlength{\abovecaptionskip}{-0.1cm}  
\setlength{\belowcaptionskip}{-0.6cm} 
\centering
\subfigure{
\includegraphics[width=0.3\linewidth]{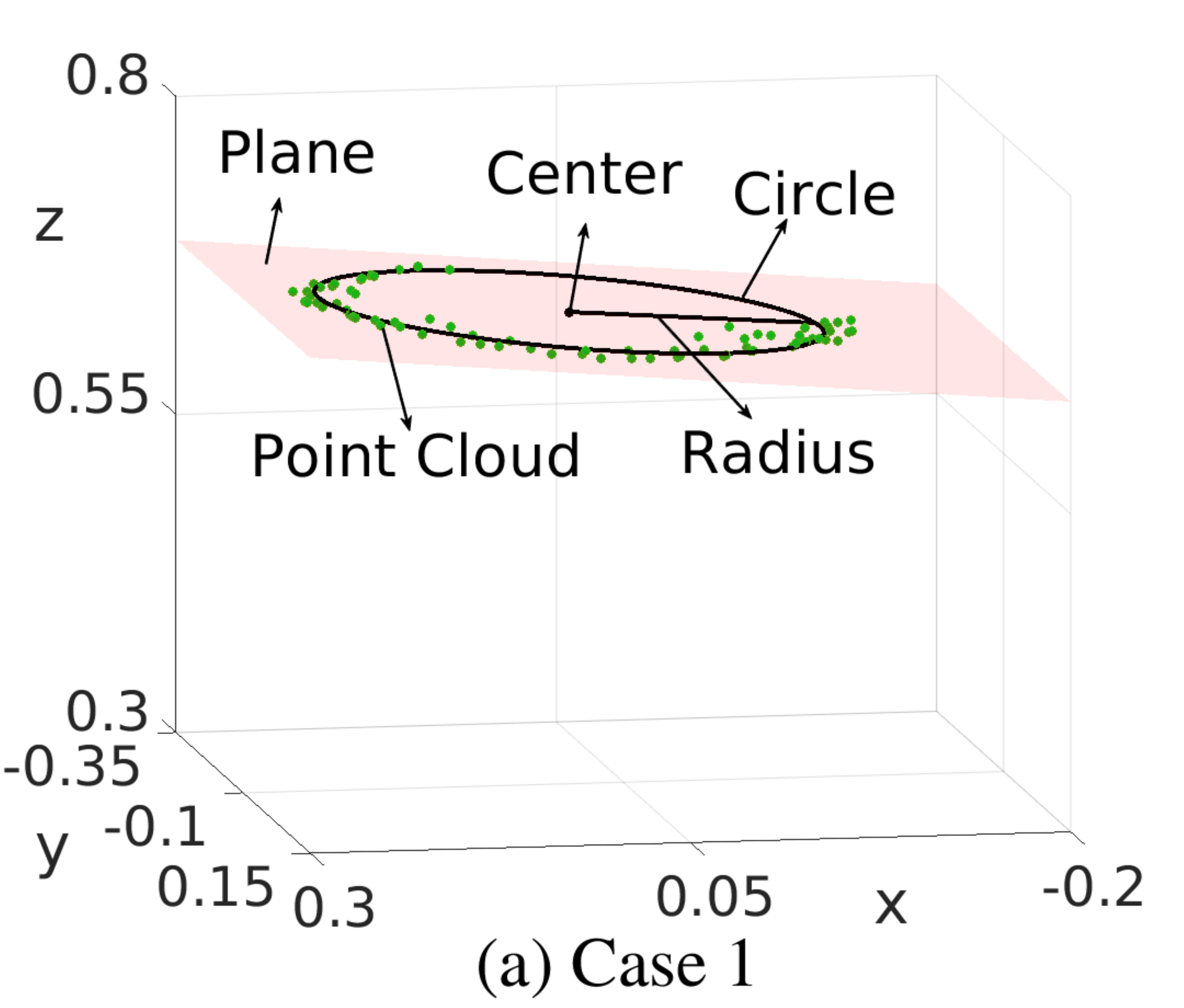}
}\ \ \ \ \ \ \ \ \ \ \ \ 
\subfigure{
\includegraphics[width=0.3\linewidth]{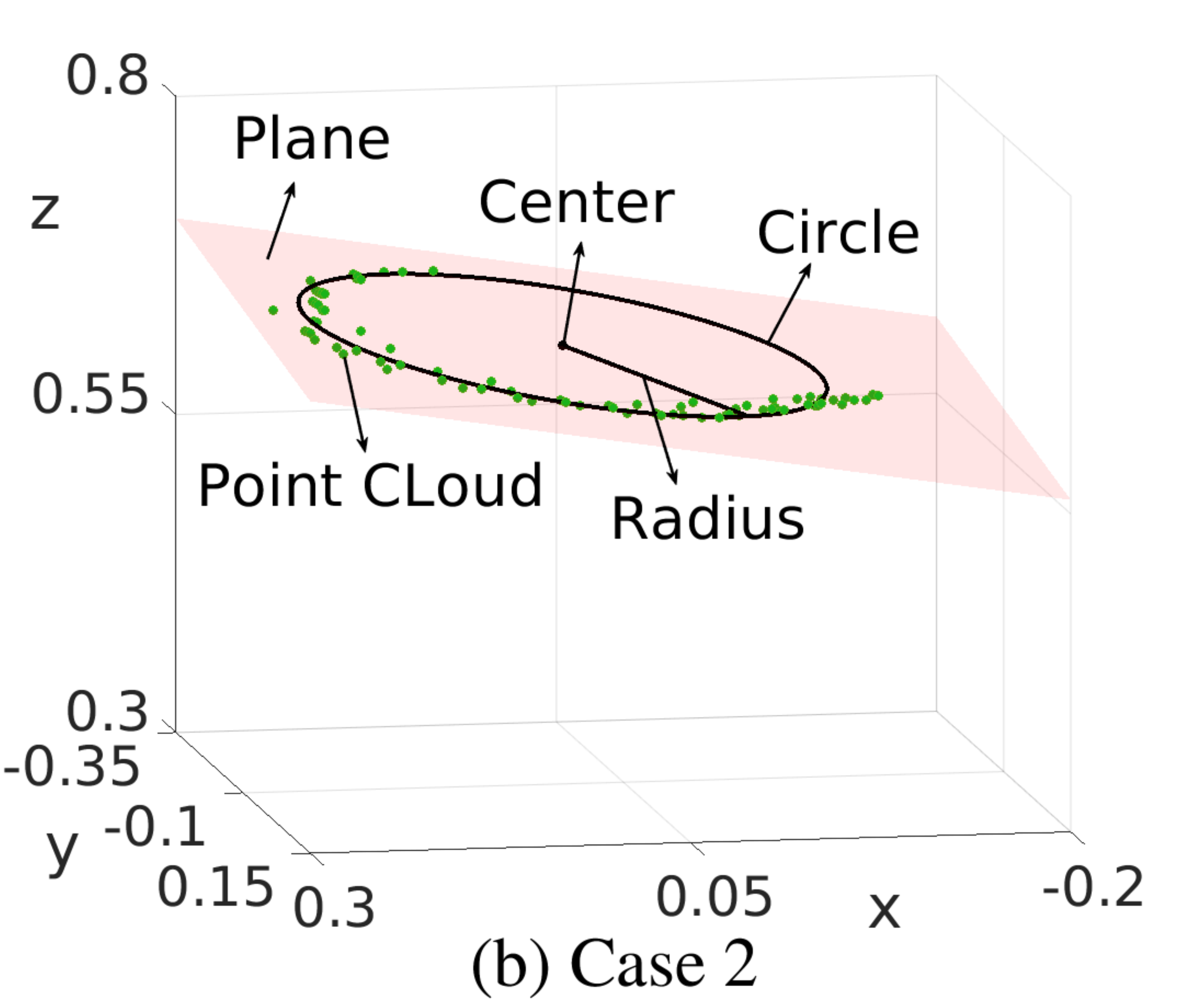}
}\\
\subfigure{
\includegraphics[width=0.3\linewidth]{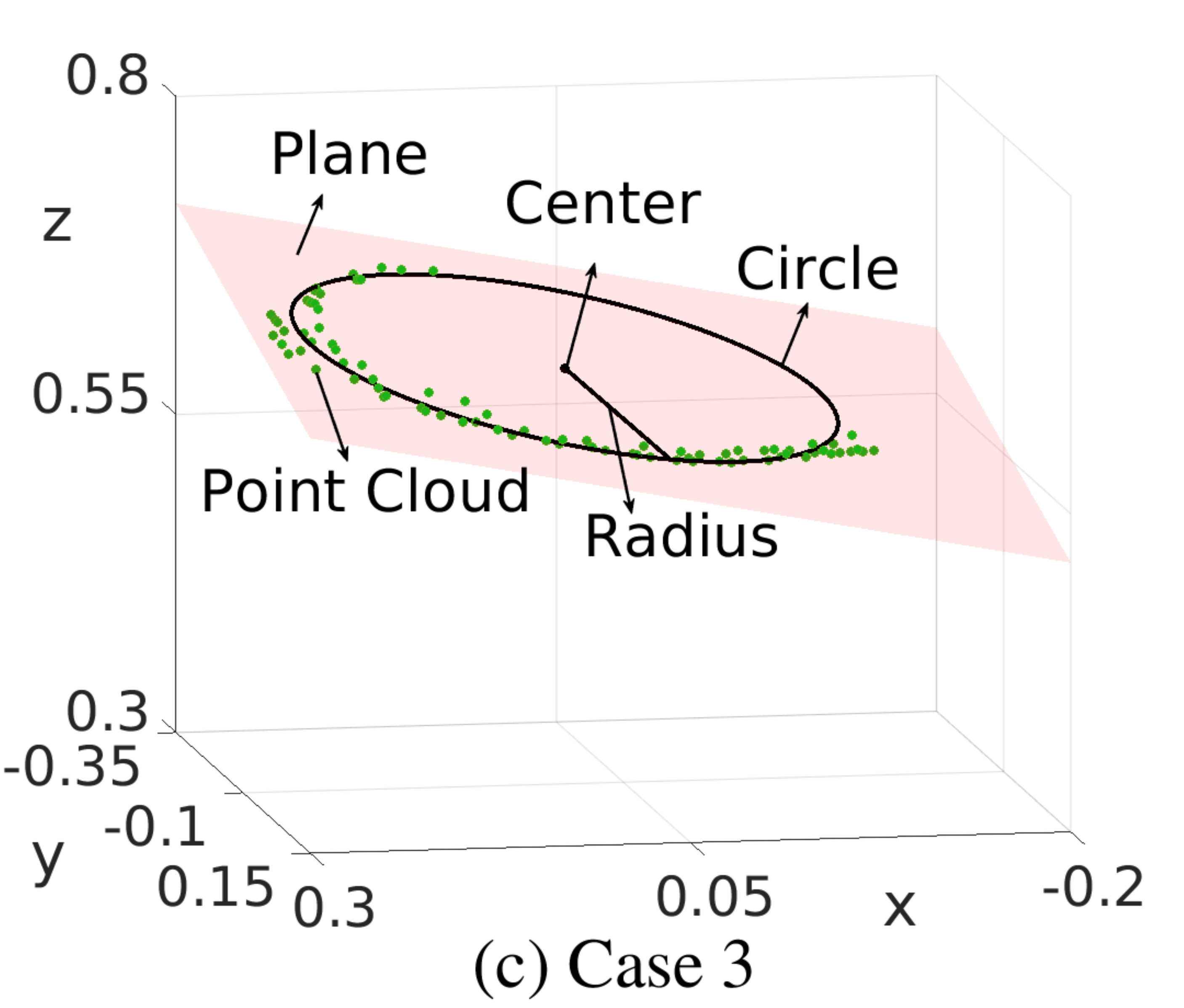}
}\ \ \ \ \ \ \ \ \ \ \ \ 
\subfigure{
\includegraphics[width=0.3\linewidth]{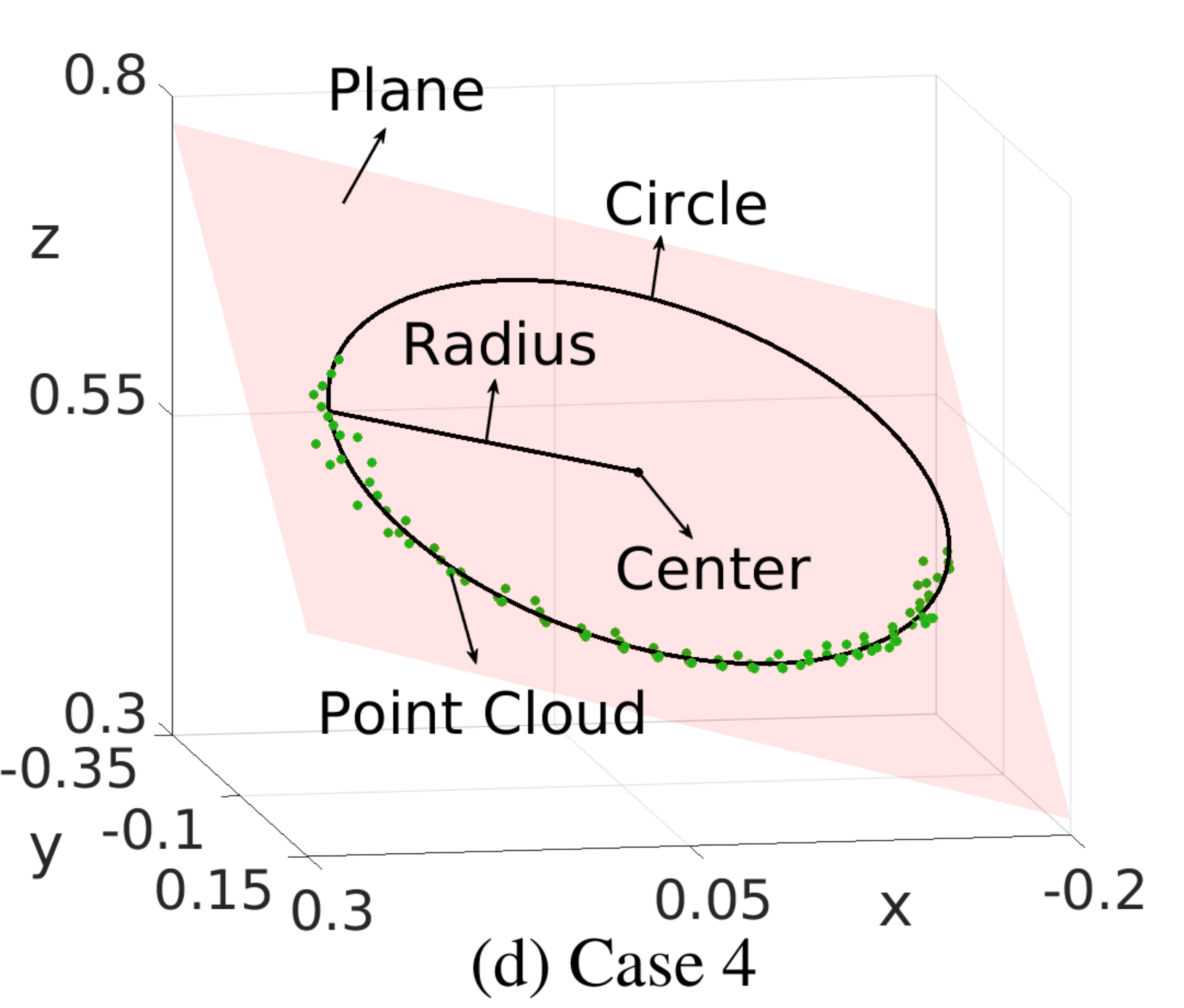}
}
\caption{The visualization of shape feature identification results.}
\label{fig:identification}
\end{figure}

The identification results are visualized as Fig.\ref{fig:identification} which are observed from the same view. 
The green points are feedback point cloud. 
The red plane represents the orientation of the normal vector.
The identified center and radius of the geometric model generate a complete circle which is drawn with a black solid line.
The regression stops around 20 ms every loop, which can achieve real-time computation.

Table \ref{table:1} shows the corresponding statistic indexes including residual mean and residual variance of the regression.
Each point's residual error is computed by $\sum^l_{i=1} f_i^2([\boldsymbol{s}^\text{T},\boldsymbol{y}^\text{T}]^\text{T})$ according to Eq.\ref{eq:rasidual}, where $\boldsymbol{s}$ is submitted by a feedback point and $\boldsymbol{y}$ is submitted by the identified shape feature.
All residual errors fall within 3 standard deviations of the mean, which indicates that residual errors follow a normal distribution according to the three-sigma rule so that the identification results are acceptable.

\begin{table}[ht!]
\vspace{-0.7cm} 
\setlength{\belowcaptionskip}{-0.6cm} 
\centering
\begin{tabular}{l| c c c c} 
 \hline
  & Case 1 & Case 2 & Case 3 & Case 4 \\ 
 \hline
 Residual Mean & 0.2385 & 0.2090 & 0.1897 & 0.0927\\ 
 Residual Variance & 0.0014 & 0.0034 & 0.0053 & 0.0045 \\
 \hline
\end{tabular}
\caption{Residual mean and variance of arc fitting results.}
\label{table:1}
\end{table}

\begin{figure}[ht!]
\vspace{-0.8cm} 
\setlength{\abovecaptionskip}{-0.2cm}  
\setlength{\belowcaptionskip}{-0.9cm} 
\begin{center}
  \includegraphics[width=0.65\linewidth]{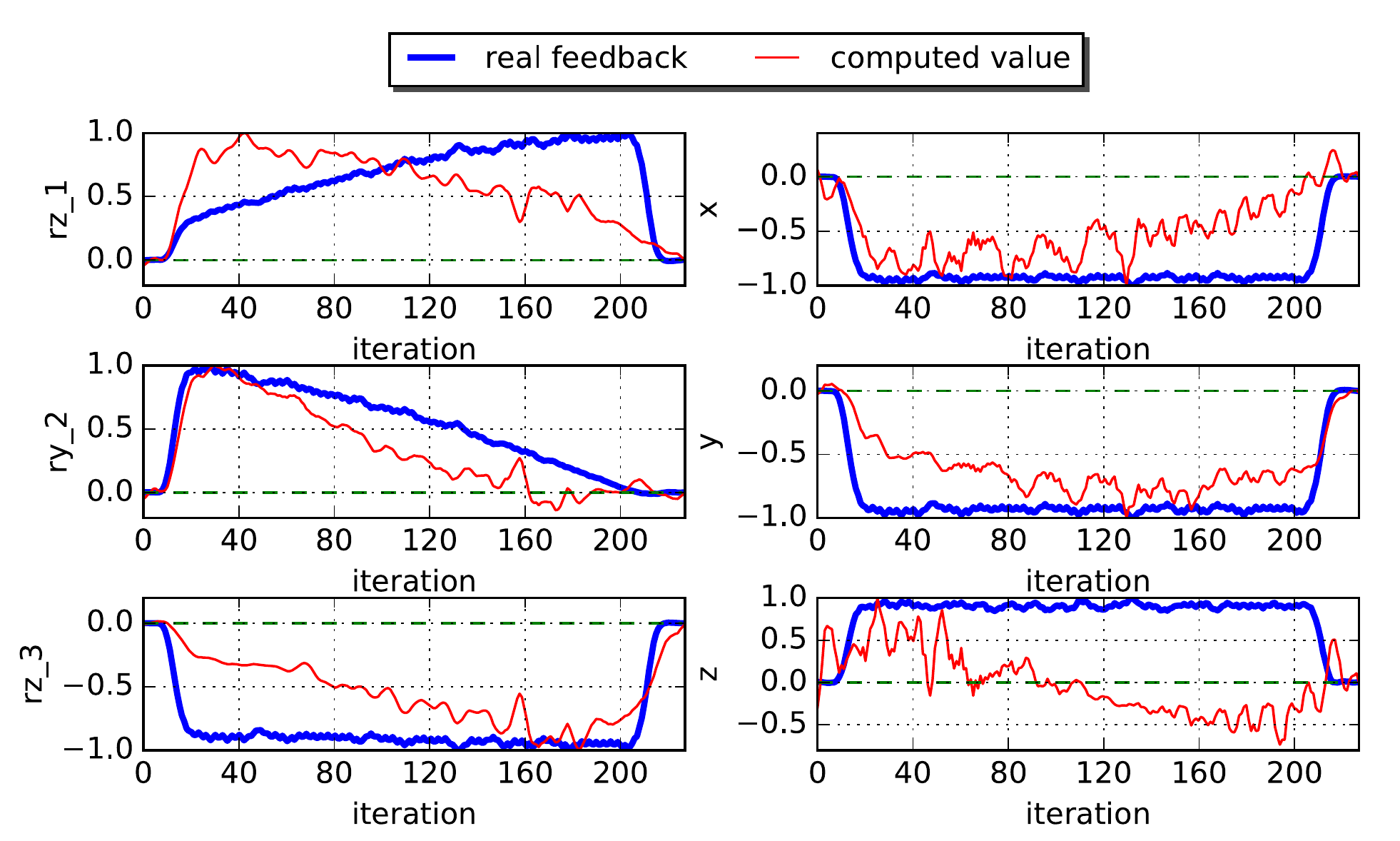}
  \caption{The comparison between computed poses and the feedback poses.}
  \label{fig:arc jacobian}    
\end{center}
\end{figure}

After the parameters identification, the Jacobian matrix should be analyzed. 
In this paper, the analytical pose-shape Jacobian matrix is evaluated.
According to Eq.\ref{eq:jacobian}, given a set of shape feature differential, the pose differential can be computed, which directs the movement of the end-effector.
Therefore, the ideal Jacobian matrix should make the computed pose differential using the feedback shape features has the same sign as the corresponding real feedback pose.
Based on this analysis, we design a movement that the robot end-effector moves along a path without control, during which process, shape features and robotic poses are collected.
Figure \ref{fig:arc jacobian} shows the comparison of the feedback poses plots(in blue thick line) and computed poses plots(in red thin line), of which, the left three subfigures display Euler angles under the rotation sequence Z-Y-Z denoted by rz\_1, ry\_2, rz\_3, respectively, and the right three subfigures display three-axial Cartesian coordinates denoted by x, y, z, respectively.
All plotted data have been normalized into the range $\left[-1,1\right]$.

Figure \ref{fig:arc jacobian} demonstrates that mostly the computed poses have the same sign as the feedback posed, which validates the proposed algorithm. 
Note that nearly half part of the z-axial coordinate has the opposite sign. 
This is because of the z-axial coordinate changes very small during the designed movement, which make the computed pose more like an oscillation around zero.
Furthermore, relatively small motion contributes little to the controller, of which the sign is not that important.
So the computed plot going through zero is reasonable.

With the proposed analytical Jacobian matrix proved to validate, the experiment that robot manipulates the soft object into the desired shape under the designed shape servoing controller is conducted.
We select all seven shape feature elements as the control target, including radius, center, and the normal vector of the spatial plane where the circle is. 
The desired shape is $\boldsymbol{y}_d=\left[0.209, -0.246, 0.359, -0.179, -0.207, 0.935, 0.289\right]^\text{T}$ and the initial state is $\boldsymbol{y}_0=\left[0.159, -0.207,  0.444, -0.178, 0.067,  0.998, -0.012\right]^\text{T}$. 

\begin{figure}[ht!]
\vspace{-0.7cm} 
\setlength{\abovecaptionskip}{-0.1cm}  
\setlength{\belowcaptionskip}{-0.9cm} 
\begin{center}
  \includegraphics[width=0.9\linewidth]{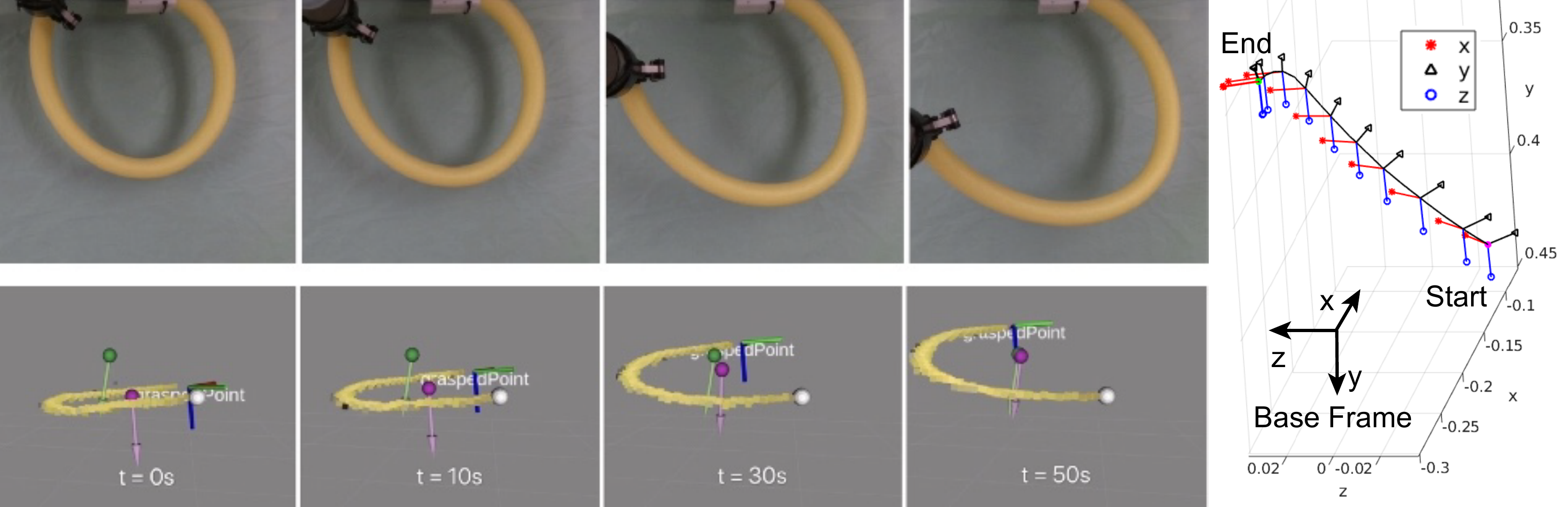}
  \caption{The control process on physical setup and simulating visualization, and the trajectories of robotic poses.}
  \label{fig:physicalprocess}    
\end{center}
\end{figure}

Figure \ref{fig:physicalprocess} demonstrates the control process (moments of t=0s, 10s, 30s, 50s are selected) on the physical setup (top view) and simulating visualization (side view).
The downside subfigures show the feedback point cloud under control. 
The white sphere is the fixed point of the deformable object.
The frame located at the other tip (grasped point) is the defined body frame.
The combination of the purple sphere and arrow represents the updating current center and normal vector.
The combination of the green sphere and arrow represents the designed target center and normal vector.
The movement that the purple combination tracks to the green combination illustrate the deformation process under the shape servoing controller.
The right-side plot in Fig. \ref{fig:physicalprocess} records the trajectory of robotic poses under the robot base frame during the experiment.
The frames along the trajectory are robotic end-effector orientation.

\begin{figure}[ht!]
\vspace{-0.8cm} 
\setlength{\abovecaptionskip}{-0.2cm}  
\setlength{\belowcaptionskip}{-0.9cm} 
\begin{center}
  \includegraphics[width=0.65\linewidth]{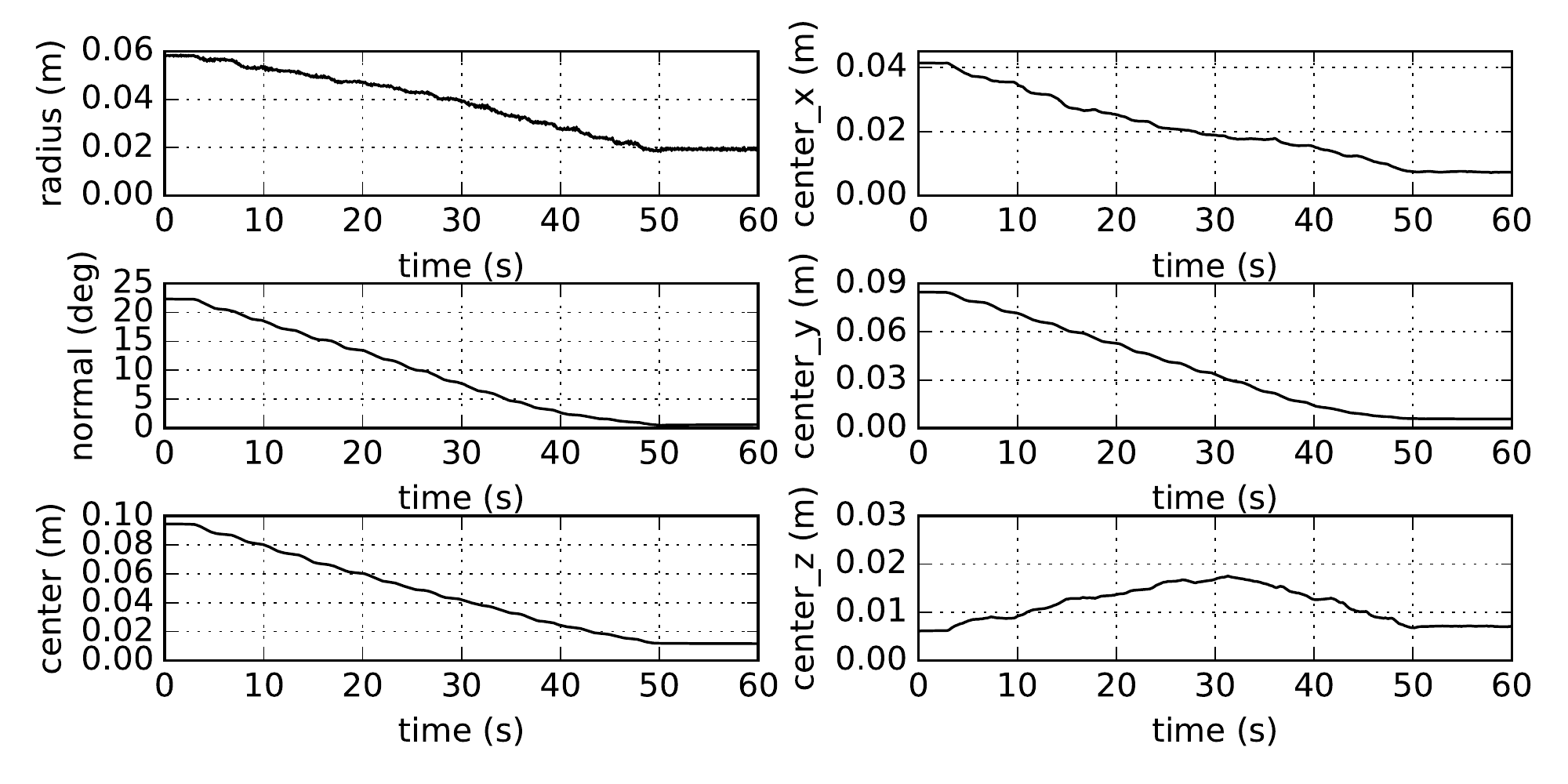}
  \caption{The visualization of shape errors.}
  \label{fig:arc shape error}    
\end{center}
\end{figure}

Figure \ref{fig:arc shape error} shows shape errors plots recorded during the movement of the robot.
The left three visualized errors are defined as $|r-r_d|$, $\arccos{ \boldsymbol{n} \cdot \boldsymbol{n}_d }$, and $\| \boldsymbol{p}_C - \boldsymbol{p}_{Cd} \|$, and the right three visualized errors are respectively three coordinates of center, defined as $|x_C - x_{Cd}|, |y_C - y_{Cd}|, |z_C - z_{Cd}|$.
It shows that shape errors converge in 50$s$.

The stable errors in Fig.\ref{fig:arc shape error} are less than 2 centimeters or degrees which is an acceptable magnitude. 
First, the point cloud has large noise while we did not use a complicated algorithm to obtain more accurate feedback data in order to reduce the calculation and achieve a real-time algorithm.   
Second, the depth camera is fixed and only obtains the point cloud from the exposed side of the manipulated object.
So that the observing view is changing along with the deformation of the elastic rod whose radius of the cross-section (about 4 cm) can lead to a bias.
Furthermore, on one hand, the defined geometric model is too ideal to approximate the whole object.
On the other hand, in our study, the deformation serves the required motion, that is, bending in this experiment, so the absolutely accurate description of shape is not necessary. 
\section{Conclusion}
In this work, we proposed a general approach to design a shape servoing controller for manipulating the deformable object into the desired shape. 
To demonstrate and validate the proposed approach, we designed a specific task of bending a elastic rod into the desired curvature with a point cloud as the sensory feedback. 
First, a continuous geometric model (spatial arc) is defined as the shape feature to globally describe deformation under bending. 
Second, use LSM to identify the parameters of the defined shape feature in real-time after filtering the raw visual feedback data. 
The statistic analysis indicates that the identification results are acceptable. 
Then, derive the analytical pose-shape Jacobian matrix based on implicit functions of the mapping of object deformation and robotic pose. 
The comparison between the computed pose and the real feedback pose during the open-loop movement of the robot proves the validation of the Jacobian matrix. 
Finally, the shape servoing controller based on velocity and the derived pose-shape Jacobian matrix enables the robot to manipulate the deformable object into the desired shape and achieve the desired motion in the task. 
The shape errors converge to acceptable stable errors. 
To sum up, both theory and experiment validate that the proposed approach can generally help design shape servoing controller based on the data-driven implicit model and achieve real-time control.

\subsubsection*{Acknowledgements.}
This work is supported in part by the Germany/Hong Kong Joint Research
Scheme sponsored by the Research Grants Council of Hong Kong and the
German Academic Exchange Service under grant G-PolyU507/18
\bibliography{biblio.bib}
\bibliographystyle{IEEEtran.bst}
\end{document}